%% file: main.tex
\documentclass[runningheads]{llncs}
\usepackage[T1]{fontenc}
\usepackage{graphicx}

\usepackage{amsmath}
\usepackage{amssymb}
\usepackage{subcaption}
\usepackage{booktabs}
\usepackage{float}
\usepackage{bbding}

\begin{document}
%
\title{GRaSp: Automatic Example Optimization for In-Context Learning in Low-Data Tasks\thanks{Code is available at \url{https://github.com/sbfroy/GRaSp}.}}
\titlerunning{GRaSp: In-Context Example Optimization for Low-Data Tasks}
%
\author{Simen Bihaug-Frøyland\inst{1}\thanks{\Envelope~\texttt{simenbih@uia.no}} \and
Henrik Brådland\inst{1,2,3}}
\authorrunning{S. Bihaug-Frøyland and H. Brådland}
%

\institute{Centre for Artificial Intelligence Research, University of Agder \and School of Computing and Information, University of Pittsburgh \and Norkart AS}
\maketitle              
\begin{abstract}
In-context learning enables large language models to adapt to new tasks, but their performance is highly sensitive to the selected examples. Finding effective demonstrations is particularly difficult in domain-specific, low-data settings where high-quality examples are scarce. We propose GRaSp, a three-stage framework for automatic in-context example optimization. By first generating a large synthetic candidate pool, then structuring it with clustering and dimensionality reduction, and finally using genetic algorithms to find the optimal in-context examples, the framework shows consistent improvements on the NER task. We also introduce a custom diversity-adaptive mutation mechanism, allowing it to transition from the initial broad inter-cluster exploration to focused intra-cluster refinement as the population converges. We evaluate GRaSp on financial named entity recognition (FiNER-139), comparing synthetic and human-annotated candidate pools across pool sizes $k \in \{500, 5000\}$. With non-synthetic data, GRaSp achieves 45.84\% $\mu$-F$_1$, consistently outperforming both zero-shot and random few-shot baselines. Synthetic data matches the random baseline but does not exceed it, suggesting that distributional variety in the candidate pool is critical for generalization.

\end{abstract}
%
%

\input{sections/introduction}
\input{sections/related_work}
\input{sections/method}
\input{sections/experiment}
\input{sections/results}
\input{sections/conclusion}

\begin{credits}

\subsubsection{\discintname}
The authors have no competing interests to declare.
\end{credits}
%
%

\bibliographystyle{splncs04}
\bibliography{references}

\end{document}

%% file: sections/introduction.tex
\section{Introduction}

Large Language Models (LLMs) have been shown to improve significantly when given carefully crafted task-specific instructions, or prompts, without performing updates to the model's parameters~\cite{sahoo_systematic_nodate}. Optimizing the instructions, known as prompt engineering, has produced several techniques for constructing effective prompts. The most prominent is In-Context Learning (ICL)~\cite{dong_survey_nodate}, where relevant examples are included in the prompt to guide the model toward the correct behavior. However, ICL relies heavily on high-quality examples as irrelevant or poorly chosen inputs have been shown to degrade performance~\cite{wu_how_2024}. 

Finding suitable examples is especially difficult for narrow, domain-specific tasks with limited data, as the available demonstrations may not cover the range of inputs encountered at inference, causing queries to fall outside the distribution of the provided examples~\cite{he_self-demos_2024}. This makes ICL less effective in low-data settings.

In our work, we therefore use LLM-generated synthetic data to build candidate pools as a first step toward optimizing example selection for few-shot prompting. Combined with clustering and Genetic Algorithms (GAs), we propose GRaSp, a three-stage framework for automatic prompt engineering in low-data tasks.

Although there is ongoing work on dynamic example selection~\cite{liu_what_2021,liu_se2_2024}, we focus exclusively on static prompts. For this work, we seek a single optimal prompt fixed before inference.

%% file: sections/related_work.tex
\section{Related Work}
\label{sec:related_work}

ICL performance is highly sensitive to the prompt's composition. Research indicates that even subtle shifts in the choice or sequence of demonstrations can swing model performance from near-random to state-of-the-art~\cite{zhao_calibrate_2021}. This section reviews contemporary efforts to mitigate this instability through automated selection, retrieval, and evolutionary optimization.

To move beyond stochastic example selection, several frameworks prioritize instance-specific retrieval. Liu et al.~\cite{liu_what_2021} demonstrated that selecting examples based on semantic proximity to the test query (KATE) significantly outperforms random sampling. Beyond similarity, information-theoretic metrics such as entropy have been employed to identify performant prompt configurations that minimize model uncertainty~\cite{lu_fantastically_2022}. While these dynamic methods offer high precision, they incur significant computational overhead at inference time. Our work instead aligns with static optimization, seeking a robust, universal set of demonstrations that performs consistently across a task domain without requiring per-query retrieval.

The arrangement of examples is as critical as their content. The $Se^2$ framework~\cite{liu_se2_2024} treats prompt construction as a sequential decision process, utilizing beam search and LLM feedback to model dependencies between examples. Similarly, PERO~\cite{kumar_reordering_2021} highlights that reordering the same set of examples can yield substantial performance gains. While PERO explicitly searches the permutation space, our proposed method treats the entire prompt sequence as a single individual within a GA. This allows the model to implicitly optimize for order and content simultaneously, favoring configurations where the interaction between order and content maximizes task accuracy.

In scenarios where high-quality human-annotated data is sparse, synthetic augmentation has emerged as a viable alternative. Recent studies have used LLMs to expand candidate pools before using GAs to prune these sets for optimal performance~\cite{balkus_improving_2024}. However, such methods often struggle with overfitting when the candidate pool is restricted. Our approach builds upon the strengths of evolutionary search but focuses on improving the search space through density-based clustering and a diversity-adaptive mutation scheme that prevents convergence on a narrow subset of examples, addressing the generalization gaps noted.

%% file: sections/method.tex
\section{Method}
\label{sec:method}
\begin{figure}[h]
    \centering
    \begin{subfigure}{0.32\textwidth}
        \centering
        \textbf{Generate}\\[2mm]
        \includegraphics[width=0.95\linewidth]{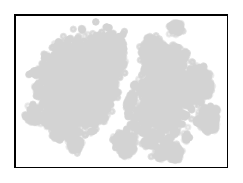}
    \end{subfigure}
    \hfill
    \begin{subfigure}{0.32\textwidth}
        \centering
        \textbf{Reduce}\\[2mm]
        \includegraphics[width=0.95\linewidth]{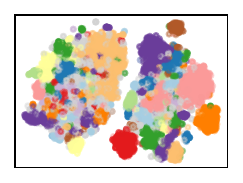}
    \end{subfigure}
    \hfill
    \begin{subfigure}{0.32\textwidth}
        \centering
        \textbf{Select}\\[2mm]
        \includegraphics[width=0.95\linewidth]{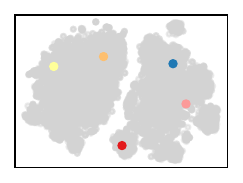}
    \end{subfigure}
    \caption{Conceptual overview of GRaSp. Generate: Produce a broad set of synthetic examples. Reduce: Cluster and filter the candidate pool. Select: Identify the most effective combination of examples by using a GA.}
    \label{fig:viz_of_grasp}
\end{figure}

We propose a three-stage approach to overcome the issue of optimizing prompt ICL-examples, termed GRaSp after its stages: Generate, Reduce, and Select. Each stage leverages established algorithms, combining the generative capabilities of LLMs with clustering and evolutionary search. GRaSp is designed to automatically identify the most effective example set for ICL within a specific task domain, enabling LLMs to adapt effectively to low-data settings. Figure~\ref{fig:viz_of_grasp} illustrates the overall process.

\subsection{Generate Stage}
At first a large candidate pool of synthetic examples, denoted by $\mathcal{S}$, is produced to cover the semantic space of the task domain. The examples are generated by an LLM prompted with an unlabeled domain corpus $D$ (e.g., PDF files). The purpose of this stage is to create a broad and diverse representation of the domain, ensuring that all relevant subtopics are captured. The focus lies on quantity and coverage rather than precision, as later stages will refine and filter the examples.

\subsection{Reduce Stage}
Then the candidate pool is structured and refined. Each example is first embedded into a vector space using a dense neural embedding model. Then, to uncover the semantic structure among examples, we employ a hierarchical density-based clustering algorithm, which groups semantically similar examples into clusters and marks outliers as noise. Removing this noise eliminates ambiguous or irrelevant examples, resulting in stable and coherent clusters that reflect the domain's main themes, forming a reduced pool $\mathcal{S}_R \subset \mathcal{S}$. This ensures that the subsequent GA focuses on optimizing between meaningful semantic patterns rather than wasting effort on misleading examples that could degrade learning quality.

\subsection{Select Stage}
At last, an evolutionary algorithm identifies the most effective combination of examples from $\mathcal{S}_R$. Each individual represents a fixed-length sequence of genes, where each gene encodes a cluster–example pair $(c,e)$: the cluster $c$ from which the example $e$ is drawn. The GA evolves the population based on a fitness function that reflects task performance (see Section~\ref{sub:test_setup} for details).

We also introduce a diversity-adaptive mutation strategy to balance the selection across versus within clusters (detailed in Section~\ref{sub:implementation_details}). Initially, the algorithm explores broadly across clusters through inter-cluster mutations to discover which semantic regions contribute most to performance. As evolution progresses, the search shifts toward intra-cluster mutation, refining examples within the most promising clusters. The process ultimately produces the optimized set $\mathcal{S}^\star \subset \mathcal{S}_R$, representing the most effective combination of examples.

By combining synthetic example generation, noise-resistant semantic clustering, and adaptive evolutionary optimization, GRaSp provides a systematic pipeline for optimizing ICL examples.

%% file: sections/experiment.tex
\section{Experimental Setup and Implementation}
\label{sec:experiment}

\subsection{Task and Data}
\label{sub:task_and_data}

The task design and label space are inspired by the FiNER-139~\cite{loukas_finer_2022} and Financial-NER-NLP datasets\footnote{\url{https://huggingface.co/datasets/Josephgflowers/Financial-NER-NLP}}, which convert XBRL annotations into natural language contexts to support entity recognition in realistic financial narratives. Following the same idea, examples are derived from two publicly available 10-K annual reports (Microsoft's FY2019 and Apple's FY2018) and expressed as self-contained, sentence-level inputs -- both sentences containing extractable financial entities and sentences without them. Extracted entity values are kept in their original surface form to retain the natural linguistic variety typical of financial reporting.

\subsection{Implementation Details}
\label{sub:implementation_details}
In the Generate stage, domain-specific examples are created from the 10-K reports. To maximize diversity, text chunks are sampled from the documents at random. The LLM \texttt{gpt-oss-120b}\footnote{An open-source large language model released by OpenAI, available at \url{https://huggingface.co/openai/gpt-oss-120b}.} generates $\mathcal{S}$ using two separate generation streams: one for positive examples (containing target entities) and one for negative examples (without target entities), ensuring balanced representation. Each stream uses dedicated prompts with explicit constraints.

In the Reduce stage, each example is first embedded using \texttt{Qwen3-Embedding-4B}\footnote{A pretrained sentence embedding model developed by Alibaba Cloud, available at \url{https://huggingface.co/Qwen/Qwen3-Embedding-4B}.} and projected into a lower-dimensional space using UMAP (Uniform Manifold Approximation and Projection). Clustering is then performed with HDBSCAN (Hierarchical Density-Based Spatial Clustering of Applications with Noise).

\begin{figure}[h]
    \centering
    \begin{subfigure}{1.0\linewidth}
        \includegraphics[width=\linewidth, trim=0.25cm 1cm 0.25cm 0cm, clip]{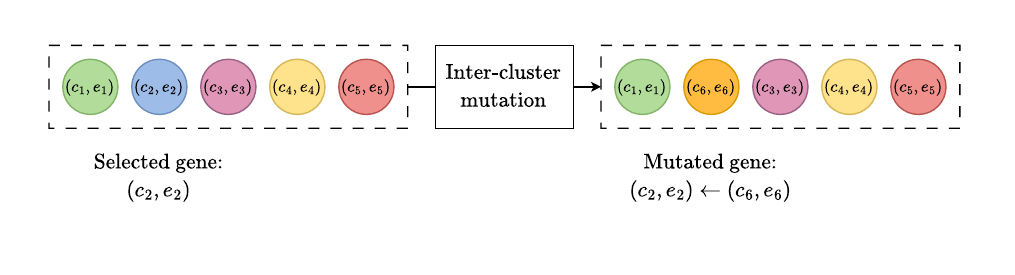}
        \subcaption{Inter-cluster mutation altering both cluster assignment and example.}
    \end{subfigure}
    \begin{subfigure}{1.0\linewidth}
        \includegraphics[width=\linewidth, trim=0.25cm 1cm 0.25cm 0cm, clip]{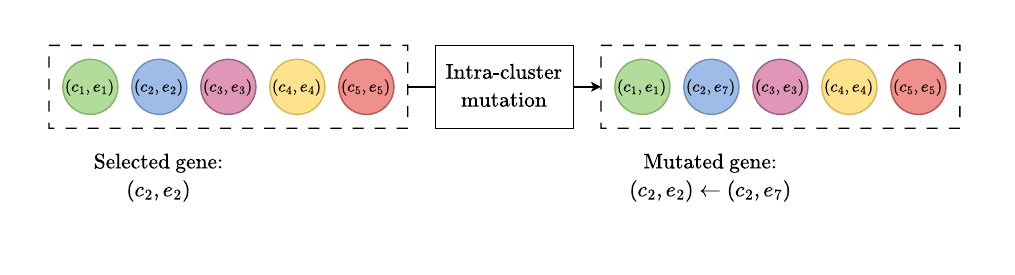}
        \subcaption{Intra-cluster mutation altering the example while preserving the cluster.}
    \end{subfigure}
    \caption{Illustration of the diversity-adaptive mutation. Colors denote cluster membership. For a given individual, one gene $(c,e)$ is selected and mutated. Inter-cluster mutation replaces both $c$ and $e$, whereas intra-cluster mutation replaces only $e$ while preserving the original cluster.}
    \label{fig:mutation}
\end{figure}

In the Select stage, a $(\mu+\lambda)$ GA -- implemented using DEAP's \texttt{eaMuPlusLambda} strategy -- is applied to derive $\mathcal{S}^\star$. It employs tournament selection, two-point crossover, and a diversity-adaptive mutation mechanism that combines inter- and intra-cluster operations (Figure~\ref{fig:mutation}). To balance exploration and exploitation, GRaSp uses a diversity-adaptive mutation probability. Population diversity $D_g$ is measured as the proportion of unique clusters represented at generation $g$. The inter-cluster mutation probability is then $p_{\text{inter}}(g) = p_{\text{min}} + (p_{\text{max}} - p_{\text{min}}) \cdot D_g$: when diversity is high, inter-cluster mutations dominate; as diversity falls, intra-cluster mutations take over.

Each individual is evaluated by measuring performance on the validation set. The LLM output is compared against gold annotations using the set-based extraction metric described in Section~\ref{sub:test_setup}.

\subsection{Hyperparameters and Evaluation Settings}
\label{sub:test_setup}
The data generation used a batch size of $20$ examples per API call. Temperature was varied across batches (base temperature $0.7 \pm 0.2$) to encourage semantic diversity. In total, 10,000 examples were generated and split into 9,000 candidate examples and 1,000 validation instances. The candidate pool was sized to ensure that sufficient examples survived noise filtering in the Reduce stage to support the largest pool size $k{=}5000$.

For the Reduce stage, the high-dimensional embeddings were first projected to $\mathbb{R}^{20}$ using UMAP. HDBSCAN was then applied with a minimum cluster size of $9$, a minimum sample size of $1$, and a cluster selection epsilon of $0.18$. For the non-synthetic data (Section~\ref{sub:experimental_p}), these parameters were re-tuned to account for differences in pool structure, using a minimum cluster size of $10$, a minimum sample size of $1$, and epsilon of $0.15$. In both cases, hyperparameters were chosen empirically to obtain semantically coherent clusters while avoiding groups that were excessively small, overly broad, or dominated by noise.

In the Select stage, the GA operated with a parent population size $\mu = 80$ and $\lambda = 180$ offspring per generation, for a maximum of $20$ generations. The crossover probability was set to $p_\text{cx} = 0.30$ and the mutation probability to $p_\text{mut} = 0.50$, with tournament selection of size $3$. The diversity-adaptive mutation scheme was configured with $p_{\text{min}} = 0.05$ and $p_{\text{max}} = 0.70$, so that high cluster diversity favored exploration via inter-cluster mutation, while reduced diversity shifted the search towards intra-cluster exploitation. Generational early stopping was applied with a patience of $5$ generations and a minimum relative improvement of $0.003$, activated after an initial warm-up of $5$ generations.

For fitness evaluation, we used an LLM run deterministically with a temperature of $0.0$. We evaluate predictions using a set-based extraction metric. For each example, the model outputs one or more labels, and for each label it outputs a set of extractable values. A predicted value counts as correct only if it exactly matches a gold value for the same label (order does not matter). We accumulate true positives, false positives, and false negatives per label across the validation set. We report the same evaluation metrics as the FiNER-139 paper: micro-precision ($\mu$-P), micro-recall ($\mu$-R), and micro-F$_1$ ($\mu$-F$_1$) computed by pooling counts across all labels (so frequent labels contribute more), and macro-F$_1$ (m-F$_1$) computed by averaging F$_1$ across labels (so each label has equal weight).

\subsection{Experimental Procedure}
\label{sub:experimental_p}
We evaluate GRaSp in two stages: (i) sensitivity to the size of $\mathcal{S}$, and (ii) comparison between synthetic and human-annotated data. All experiments follow the pipeline described in Sections~\ref{sub:task_and_data}--\ref{sub:test_setup}.

To study the impact of pool size, we construct two candidate pools with $k \in \{500, 5000\}$ in-context examples. We first run the Generate and Reduce stages to obtain a clustered, noise-filtered set $\mathcal{S}_R$. Then, for each value of $k$, we build $\mathcal{S}_k$ via round-robin sampling over clusters: examples are drawn one at a time from each cluster in turn, cycling from the largest to the smallest until $k$ examples have been selected. This ensures that all clusters are represented as uniformly as possible.

We also apply the same procedure directly to the original Financial-NER-NLP dataset. In these cases, the labeled Financial-NER-NLP examples form the initial candidate pool $\mathcal{S}$, and the same clustering, sampling, and evolutionary procedure is repeated. This allows us to assess GRaSp under near-ideal conditions using non-synthetic data.

Reduced subsets are used during evolutionary optimization due to the large size of the available training and validation data. The subsets are sampled to preserve the original label distribution and to align with the size of the generated synthetic data. Specifically, $9,000$ training instances are used instead of $\sim900,000$, and $1,000$ validation instances instead of $\sim100,000$.

%% file: sections/results.tex
\section{Results and Discussion}
\label{sec:results}

\begin{figure}[h]
    \centering
    \begin{subfigure}{0.48\linewidth}
        \centering
        \includegraphics[width=\linewidth]{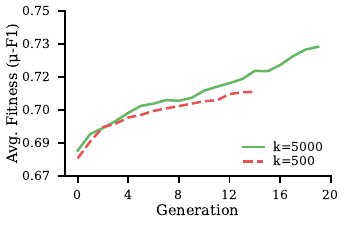}
        \caption{Synthetic data}
        \label{fig:fitness_synthetic}
    \end{subfigure}
    \hfill
    \begin{subfigure}{0.48\linewidth}
        \centering
        \includegraphics[width=\linewidth]{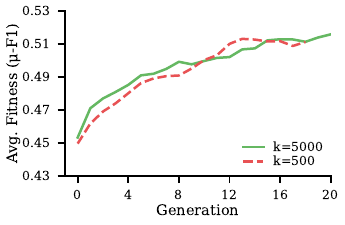}
        \caption{Non-synthetic data}
        \label{fig:fitness_non_synthetic}
    \end{subfigure}
    \caption{Average validation fitness over generations for different pool sizes. Each curve shows the mean fitness ($\mu$-F$_1$) across the population at each generation.}
    \label{fig:fitness_over_gens}
\end{figure}

Figure~\ref{fig:fitness_over_gens} shows the average population fitness on the validation set across generations. Note that the synthetic and non-synthetic configurations use separate validation sets: the synthetic runs are evaluated against LLM-generated validation data, while the non-synthetic runs use human-annotated examples. Fitness scores are therefore not directly comparable across the two settings. All configurations show steady improvement, confirming that the GA effectively optimizes example selection. The synthetic runs start at a notably higher fitness ($\sim$ 0.68 $\mu$-F$_1$) than the non-synthetic ones ($\sim$ 0.45), with $k{=}5000$ reaching $\sim$0.73 by generation 20. This higher starting point likely reflects that the synthetic validation set was generated by the same LLM as the candidate pool $\mathcal{S}$, making it easier for any example set to score well. The non-synthetic runs, drawn from human-annotated data with greater linguistic and factual variety, start lower but show consistent gains across all pool sizes, with $k{=}5000$ reaching the highest final fitness ($\sim$ 0.52).
\begin{table}[t]
\centering
\caption{Test set results for ICL methods. Random baselines report mean $\pm$ std over $3$ independent draws. GRaSp configurations use five in-context examples selected from candidate pools of size $k$. Bold indicates best result per column.}
\label{tab:test_results}
\vspace{6pt}
\small
\setlength{\tabcolsep}{6pt}
\renewcommand{\arraystretch}{1.15}
\begin{tabular}{llcccc}
\toprule
& & $\mu$-P (\%) & $\mu$-R (\%) & $\mu$-F$_1$ (\%) & m-F$_1$ (\%) \\
\midrule
& Zero-shot & 30.64 & 58.58 & 40.23 & 33.22 \\
\midrule
\multicolumn{6}{l}{\textit{Synthetic data}} \\
& Random              & 35.14 $\pm$ 0.43 & 57.11 $\pm$ 0.58 & 43.51 $\pm$ 0.48 & 36.38 $\pm$ 0.20 \\
& GRaSp ($k{=}5000$)  & \textbf{39.53} & 47.62 & 43.20 & 35.55 \\
& GRaSp ($k{=}500$)   & 36.69 & 52.84 & 43.31 & 36.58 \\
\midrule
\multicolumn{6}{l}{\textit{Non-synthetic data}} \\
& Random              & 33.93 $\pm$ 0.86 & 58.40 $\pm$ 1.80 & 42.91 $\pm$ 1.06 & 36.43 $\pm$ 1.22 \\
& GRaSp ($k{=}5000$)  & 36.96 & 60.32 & \textbf{45.84} & \textbf{39.18} \\
& GRaSp ($k{=}500$)   & 36.37 & \textbf{61.01} & 45.57 & 38.86 \\
\bottomrule
\end{tabular}
\end{table}

As seen in Table~\ref{tab:test_results}, all few-shot methods improve over zero-shot, which achieves high recall ($58.58\%$) but low precision ($30.64\%$), yielding a $\mu$-F$_1$ of $40.23\%$. Among non-synthetic configurations, GRaSp consistently outperforms the random baseline across all metrics. The best result, $\mu$-F$_1$ = $45.84\%$ with $k{=}5000$, represents a $2.93$ percentage point (pp) improvement over the random baseline ($42.91\%$) and a $5.61$ point gain over zero-shot. Performance is remarkably stable across pool sizes. This stability suggests that the Reduce stage plays an important role: by clustering $\mathcal{S}$ and sampling across clusters, GRaSp ensures that even small pools cover the semantic space of the task, giving the GA a well-structured search space regardless of $k$.

The synthetic results tell a different story. GRaSp ($k{=}500$) and GRaSp ($k{=}5000$) roughly match the random baseline on $\mu$-F$_1$ ($\sim 43\%$) but do not exceed it. Notably, GRaSp ($k{=}5000$) achieves the highest precision overall ($39.53\%$) at the cost of lower recall ($47.62\%$), indicating that the GA converged on a conservative example set that favors precision.

The contrast between synthetic and non-synthetic results is noticeable. Despite the higher validation fitness observed for synthetic data (Figure~\ref{fig:fitness_over_gens}), this advantage does not transfer to the test set. The synthetic examples, while sufficient for the GA to optimize against its own validation data, appear to lack the distributional variety needed to produce prompts that generalize to real text.

\begin{figure}[h]
    \centering
    \begin{subfigure}{0.48\linewidth}
        \centering
        \includegraphics[width=\linewidth]{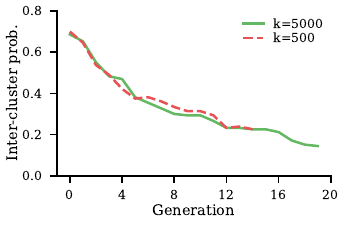}
        \caption{Synthetic data}
        \label{fig:mutation_synthetic}
    \end{subfigure}
    \hfill
    \begin{subfigure}{0.48\linewidth}
        \centering
        \includegraphics[width=\linewidth]{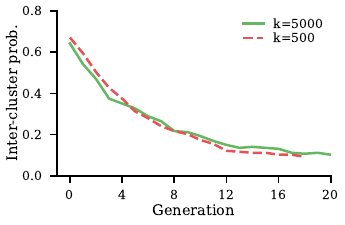}
        \caption{Non-synthetic data}
        \label{fig:mutation_non_synthetic}
    \end{subfigure}
    \caption{Adaptive inter-cluster mutation probability over generations for different pool sizes. Each curve shows the inter-cluster mutation probability at each generation.}
    \label{fig:mutation_prob}
\end{figure}

Figure~\ref{fig:mutation_prob} visualizes the diversity-adaptive mutation mechanism. In all configurations, the inter-cluster mutation probability starts high and decays over the course of evolution, confirming that the mechanism operates as intended: the population begins with broad exploration across clusters and gradually shifts toward intra-cluster refinement as the GA converges on the most promising semantic regions. The decay is steepest for $k{=}5000$, where the larger pool offers more initial diversity that is progressively narrowed. This validates the core design premise of the adaptive mutation, namely that the exploration--exploitation balance is driven by the population itself rather than by a fixed schedule.

The non-synthetic results demonstrate that GRaSp's optimization does generalize to unseen data: both pool sizes consistently outperform the random baseline on the held-out test set (Table~\ref{tab:test_results}). This is notable because prior work on GA-based prompt optimization has reported poor generalization~\cite{balkus_improving_2024}. Balkus and Yan used a GA to select in-context examples for GPT-3 and achieved $85\%$ validation accuracy but only $67\%$ on unseen data, attributing the gap to overfitting from a small candidate pool. GRaSp's clustering and noise filtering mitigate this by structuring the search space into semantically coherent clusters, allowing the GA to select examples that cover the task domain rather than overfitting to idiosyncratic validation patterns.

The synthetic configurations, however, do not share this advantage. Despite higher validation fitness (Figure~\ref{fig:fitness_over_gens}), the GA-optimized synthetic prompts fail to outperform random selection on the test set. The issue is not overfitting in the traditional sense -- the GA operates correctly and fitness improves -- but rather a distributional mismatch between the synthetic $\mathcal{S}$ and the real evaluation data. The LLM-generated examples likely capture frequent patterns in the domain but underrepresent the linguistic and factual variation found in actual financial 10-K reports. The high synthetic validation fitness further supports this interpretation as the synthetic validation set, generated by the same process, is easy for any synthetic prompt to score well on, but this does not indicate genuine task competence on real financial texts. These results suggest that GRaSp's clustering and evolutionary optimization are effective when the candidate pool is representative of the target domain, but that the framework's performance is ultimately bounded by the quality of the data it operates on.
\begin{table}[t]
\centering
\caption{Comparison to supervised baselines fine-tuned on the full training set (mean $\pm$ std over 3 seeds). GRaSp reports the best-performing configuration (non-synthetic, $k{=}5000$).}
\label{tab:supervised_baselines}
\vspace{6pt}
\small
\setlength{\tabcolsep}{6pt}
\renewcommand{\arraystretch}{1.15}
\begin{tabular}{lcc}
\toprule
& $\mu$-F$_1$ (\%) & m-F$_1$ (\%) \\
\midrule
spaCy    & 48.6 $\pm$ 0.4 & 37.6 $\pm$ 0.2 \\
BiLSTM   & 71.3 $\pm$ 0.2 & 68.6 $\pm$ 0.2 \\
BERT     & 75.1 $\pm$ 1.1 & 72.6 $\pm$ 1.4 \\
FinBERT  & 74.0 $\pm$ 1.1 & -- \\
SEC-BERT & 75.7 $\pm$ 0.1 & 72.6 $\pm$ 0.4 \\
\midrule
GRaSp    & 45.84 & 39.18 \\
\bottomrule
\end{tabular}
\end{table}

For completeness, Table~\ref{tab:supervised_baselines} compares GRaSp to supervised baselines from the FiNER-139 benchmark. These models are fine-tuned on the full training set ($\sim$900k examples) and, as expected, substantially outperform GRaSp. The gap underscores the difficulty of the task: representing over 139 XBRL tags is inherently challenging for ICL with only five examples. Nevertheless, GRaSp operates in a fundamentally different regime -- no parameter updates, no labeled training data beyond the few-shot prompt -- and its best configuration approaches the performance of spaCy (48.6\% $\mu$-F$_1$), which itself uses supervised training.

%% file: sections/conclusion.tex
\section{Conclusion}
\label{sec:conclusion}

In this paper, we present GRaSp, a three-stage framework for automatic in-context example optimization in low-data tasks. By combining synthetic data generation, density-based clustering, and a GA with diversity-adaptive mutation, GRaSp provides a systematic approach to constructing effective few-shot prompts without requiring labeled training data. Applied to financial NER on the FiNER-139 task, GRaSp with non-synthetic data achieves $45.84\%$ $\mu$-F$_1$, improving over both zero-shot (+$5.61$ pp) and random few-shot baselines (+$2.93$ pp). The adaptive mutation mechanism successfully transitions from exploration to exploitation as the population converges, as confirmed by the decay in inter-cluster mutation probability across generations.

Our results also reveal some important limitations. While GRaSp generalizes well with non-synthetic data, synthetic candidate pools do not yield gains over random selection, pointing to data quality as the binding constraint. Furthermore, the absolute performance gap to supervised baselines remains large, reflecting the difficulty of 139-class NER with only five in-context examples.

Several directions remain for future work. Evaluating GRaSp on additional tasks and domains would test the generality of the framework. Improving synthetic data quality -- for instance through filtering, iterative refinement, or incorporating real examples as seeds during generation -- is perhaps the most pressing direction, as closing the gap between synthetic and non-synthetic performance would make GRaSp fully viable in tasks where no labeled data exists at all. Finally, scaling to larger candidate pools and longer evolutionary runs may yield further gains, as most configurations did not trigger early stopping within the allocated budget.